\documentclass[letterpaper]{article} 
\usepackage{aaai2026}  
\usepackage{times}  
\usepackage{helvet}  
\usepackage{courier}  
\usepackage[hyphens]{url}  
\usepackage{graphicx} 
\usepackage{natbib}  
\usepackage{caption} 
\usepackage{algorithm}
\usepackage{algorithmic}

\usepackage{multirow}
\usepackage{adjustbox}
\usepackage{listings}
\usepackage{amsmath}
\usepackage{amsthm}
\usepackage{algorithm}
\usepackage{algorithmic}
\usepackage{graphicx}
\usepackage{bm}
\usepackage{booktabs}
\usepackage{amssymb}
\usepackage{subfig}
\usepackage{newfloat}
\usepackage{listings}
\DeclareCaptionStyle{ruled}{labelfont=normalfont,labelsep=colon,strut=off} 
\floatstyle{ruled}
\newfloat{listing}{tb}{lst}{}
\floatname{listing}{Listing}

\title{Decoupling Template Bias in CLIP: Harnessing Empty Prompts for Enhanced Few-Shot Learning}
\author{
    Zhenyu Zhang\textsuperscript{\rm 1},
    Guangyao Chen\textsuperscript{\rm 2}\footnotemark[1],
    Yixiong Zou\textsuperscript{\rm 1}\thanks{Corresponding authors.},
    Zhimeng Huang\textsuperscript{\rm 2},
    Yuhua Li\textsuperscript{\rm 1}\footnotemark[1]
}
\affiliations{
    \textsuperscript{\rm 1}School of Computer Science and Technology, Huazhong University of Science and Technology\\
    \textsuperscript{\rm 2}National Key Laboratory for Multimedia Information Processing, School of Computer Science, Peking University\\

}
\newtheorem{theorem}{Theorem}[section]
\newtheorem{observation}[theorem]{Observation}

\usepackage{bibentry}

\begin{document}

\maketitle

\begin{abstract}

The Contrastive Language-Image Pre-Training (CLIP) model excels in few-shot learning by aligning visual and textual representations. Our study shows that template-sample similarity (TSS), defined as the resemblance between a text template and an image sample, introduces bias. This bias leads the model to rely on template proximity rather than true sample-to-category alignment, reducing both accuracy and robustness in classification.
We present a framework that uses empty prompts, textual inputs that convey the idea of “emptiness” without category information. These prompts capture unbiased template features and offset TSS bias. The framework employs two stages. During pre-training, empty prompts reveal and reduce template-induced bias within the CLIP encoder. During few-shot fine-tuning, a bias calibration loss enforces correct alignment between images and their categories, ensuring the model focuses on relevant visual cues.
Experiments across multiple benchmarks demonstrate that our template correction method significantly reduces performance fluctuations caused by TSS, yielding higher classification accuracy and stronger robustness. The repository of this project is available at \url{https://github.com/zhenyuZ-HUST/Decoupling-Template-Bias-in-CLIP}.

\end{abstract}    

\section{Introduction}
\label{sec:intro}
CLIP (Contrastive Language-Image Pre-Training)~\cite{radford2021learning} is a multimodal pre-trained neural network designed to align images and text using large-scale paired image-text data. The model consists of two branches: a text encoder and an image encoder, each mapping textual descriptions and visual samples into low-dimensional vector representations. During pre-training, CLIP learns to perform a wide range of tasks, including OCR~\cite{materzynska2022disentangling}, geolocation~\cite{vivanco2024geoclip}, and action recognition~\cite{ke2018learning}. In the prediction phase, CLIP generates predictions by calculating the cosine similarity between text and image vectors. This model is particularly effective for zero-shot learning tasks, where it can make predictions without having seen training examples of new images or text.


\begin{figure*}[!t]
\centering
\includegraphics[width=1\linewidth]{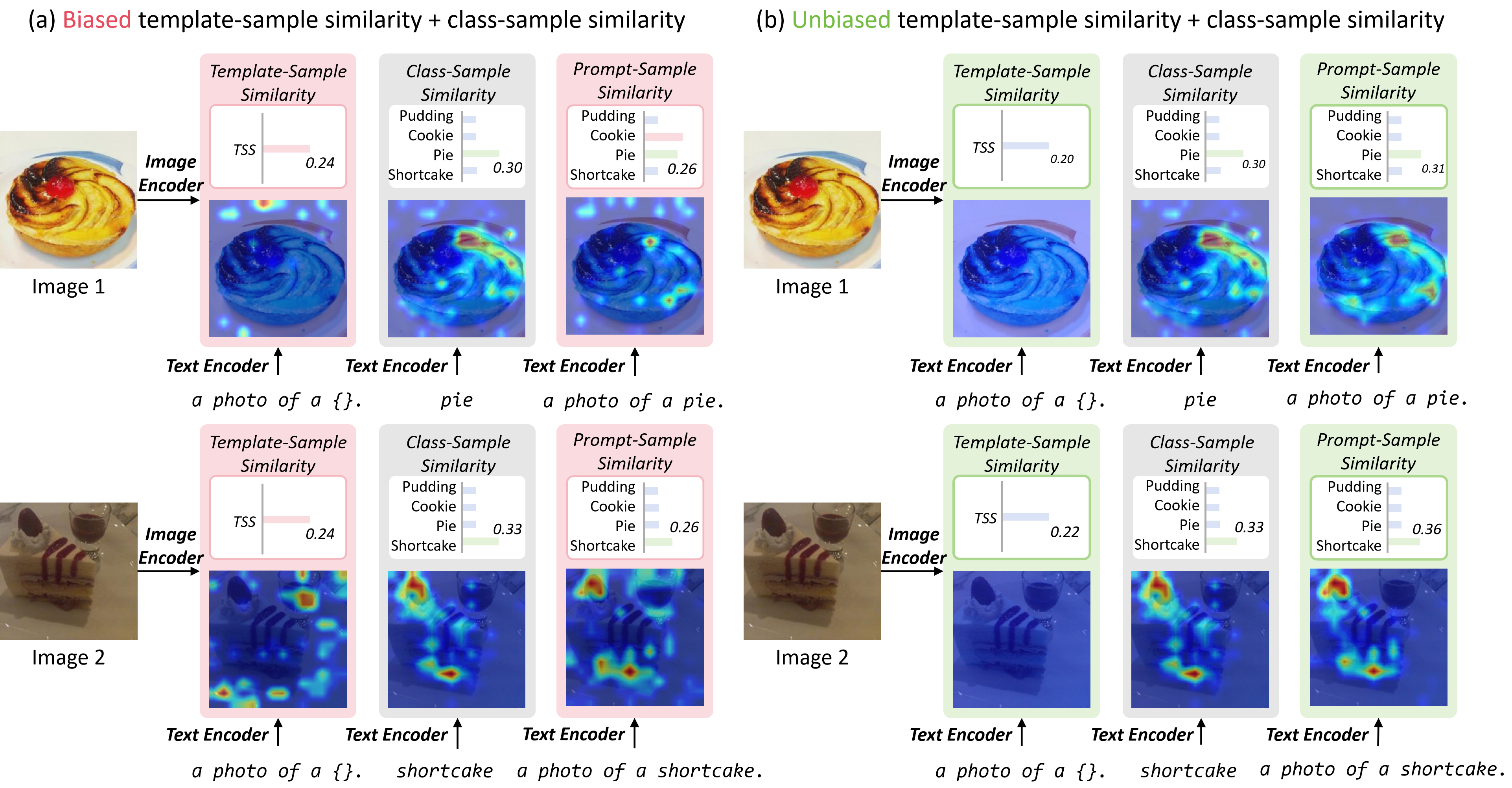}
\caption{Impact of template bias on CLIP classification. (a) A biased prompt (“a photo of a {}”) introduces extra template–sample similarity that skews attention toward irrelevant regions and inflates incorrect class scores—even when the class name matches the object. (b) An unbiased prompt removes template bias, so the model focuses on genuinely discriminative features and the model’s prediction relies solely on class–sample similarity and correct prompt–sample similarity.
.} 
\label{fig:finding}
\end{figure*}


In downstream few-shot learning tasks, CLIP classifies an image by encoding text descriptions such as “a photo of a cat” and comparing the resulting embeddings to image features. We define three cosine‐based similarity measures: \emph{template‐sample similarity} (TSS), between an image and the embedding of the blank template ``a photo of a \{\}''; \emph{class‐sample similarity} (CSS), between an image and the embedding of the bare class name (e.g., “cat”); and \emph{prompt‐sample similarity} (PSS), between an image and the embedding of the full prompt (e.g., “a photo of a cat”). 
Prior work has investigated the impact of classname variations on CLIP’s zero‐shot performance~\cite{zhang2022contrastive,zhu2023generalized,shinotter,cabello2023evaluating} and demonstrated that carefully engineered prompts can yield substantial accuracy gains~\cite{yang2023mitigating,you2024calibrating,wen2024makes,chen2024catastrophic,fabbrizzi2022survey,seth2023dear}.  In parallel, several studies have shown that different template choices can induce large fluctuations in performance~\cite{liusie2023mitigating,he2024does}.  As illustrated in the upper panel of Figure~\ref{fig:finding}, encoding each image with the generic template “\texttt{a photo of a \{\}}” produces systematically varying Template–Sample Similarity (TSS) scores, and appending the true class name yields different Prompt–Sample Similarity (PSS) values.  Concretely, images with higher TSS scores consistently generate higher PSS across all class prompts. Most importantly, \textbf{the model exhibits an attention bias even when the input is a blank template, and this bias persists with a full prompt (including category information), resulting in incorrect final predictions}.  These findings pose an important question: \textit{although templates supply crucial contextual cues, might they also introduce systematic biases that limit CLIP’s ability to discriminate among classes and thereby constrain its overall effectiveness?}

To elucidate the effects of template–sample similarity (TSS) and optimization strategies on few-shot performance, we performed a detailed analysis of how different prompt templates influence classification accuracy. We find that, in low-data regimes, there is a strong correlation between an image’s TSS score and its likelihood of being correctly classified, suggesting that models often exploit template similarity instead of truly aligning samples with their correct labels. Existing few-shot methods largely ignore this template-induced distortion. When only a handful of examples are available, such biases cannot be overcome by data augmentation or extensive training alone. Consequently, explicitly accounting for TSS is essential to improve both the robustness and the accuracy of few-shot learning systems.

Based on these findings, we propose a method to eliminate the bias caused by the relationship between templates and samples, thereby further improving few-shot learning. As shown in Figure~\ref{fig:our_method}, we first construct multiple empty prompts by combining the template with words that represent “emptiness”. This captures the feature representations of the template's textual space when category information is absent. We address the template bias by ensuring that these empty prompts exhibit consistent similarity with the samples, forcing the template-sample similarity to be uniform across all samples and correcting the bias introduced by the template. Second, during the pretraining phase, these empty prompts are used to detect and correct biases within the CLIP model, ensuring that the initial model parameters are less influenced by template similarities. Finally, we perform few-shot fine-tuning using limited labeled samples, simultaneously applying a bias calibration loss to maintain the model's focus on sample-category alignment rather than template similarity. This two-stage training approach effectively reduces the dependency on TSS, leading to enhanced classification accuracy and robustness across various templates and categories. Our method ensures that the model relies more on the intrinsic alignment between samples and their true categories, thereby improving the fairness and performance of few-shot learning models.

In summary, our contributions are as follows: 
\begin{itemize} 
\item We uncover and analyze significant performance fluctuations in CLIP-based models caused by the similarity between text templates and image samples. This indicates that while templates improve the overall performance, they also introduce a non-negligible prior bias.

\item We construct multiple “empty” prompts representing the concept of “emptiness” capturing the textual feature representations of templates without category information. 
\item Based on these insights, we introduce a two-stage training strategy that effectively reduces dependency on template-sample similarity, enhancing accuracy and robustness in few-shot learning. 
\item Extensive experiments on multiple datasets demonstrate that our template correction method significantly reduces performance fluctuations caused by templates and improves the overall performance of CLIP-based few-shot learning models. 
\end{itemize}

\section{Template-Sample Similarity}
\label{sec:tss}

\begin{figure*}[h]
\centering
\begin{minipage}{.36\linewidth}
    \centering
    \subfloat{
    \includegraphics[width=1\linewidth]{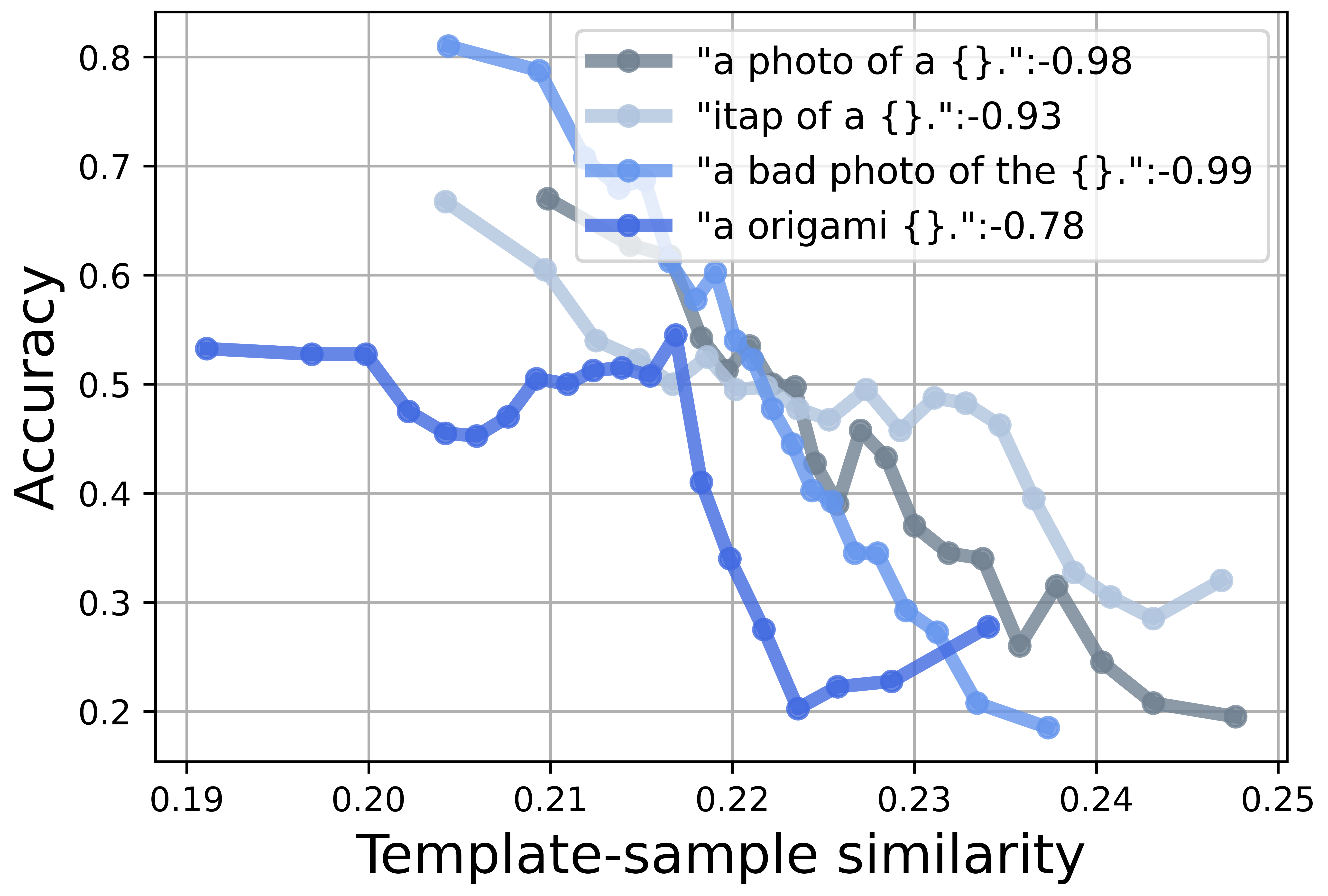}
    }
\end{minipage}
\medskip
\begin{minipage}{.26\linewidth}
    \centering
    \subfloat{
    \includegraphics[width=1\linewidth]{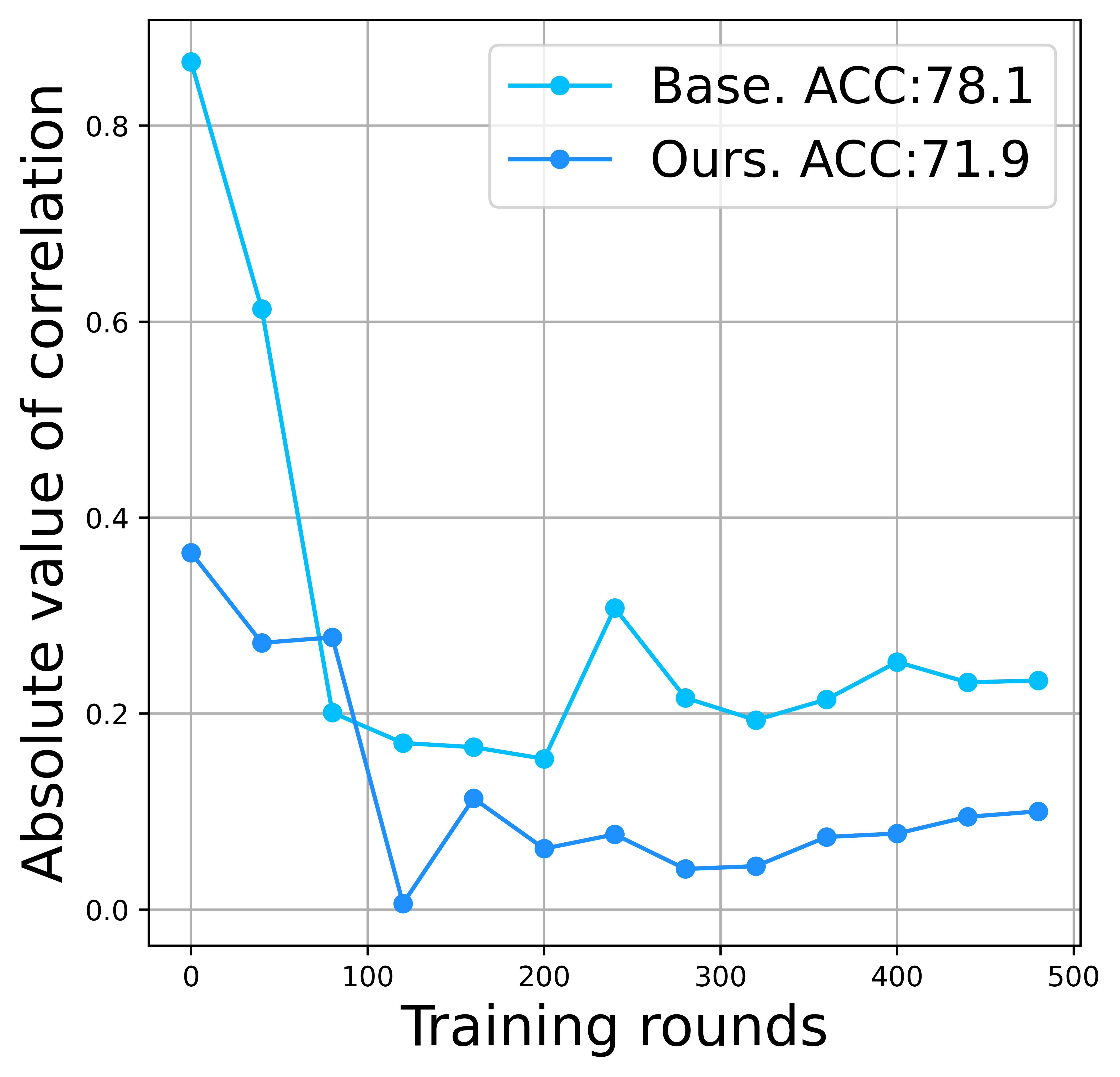}
    }
\end{minipage}
\begin{minipage}{.36\linewidth}
    \centering
    \subfloat{
    \includegraphics[width=1\linewidth]{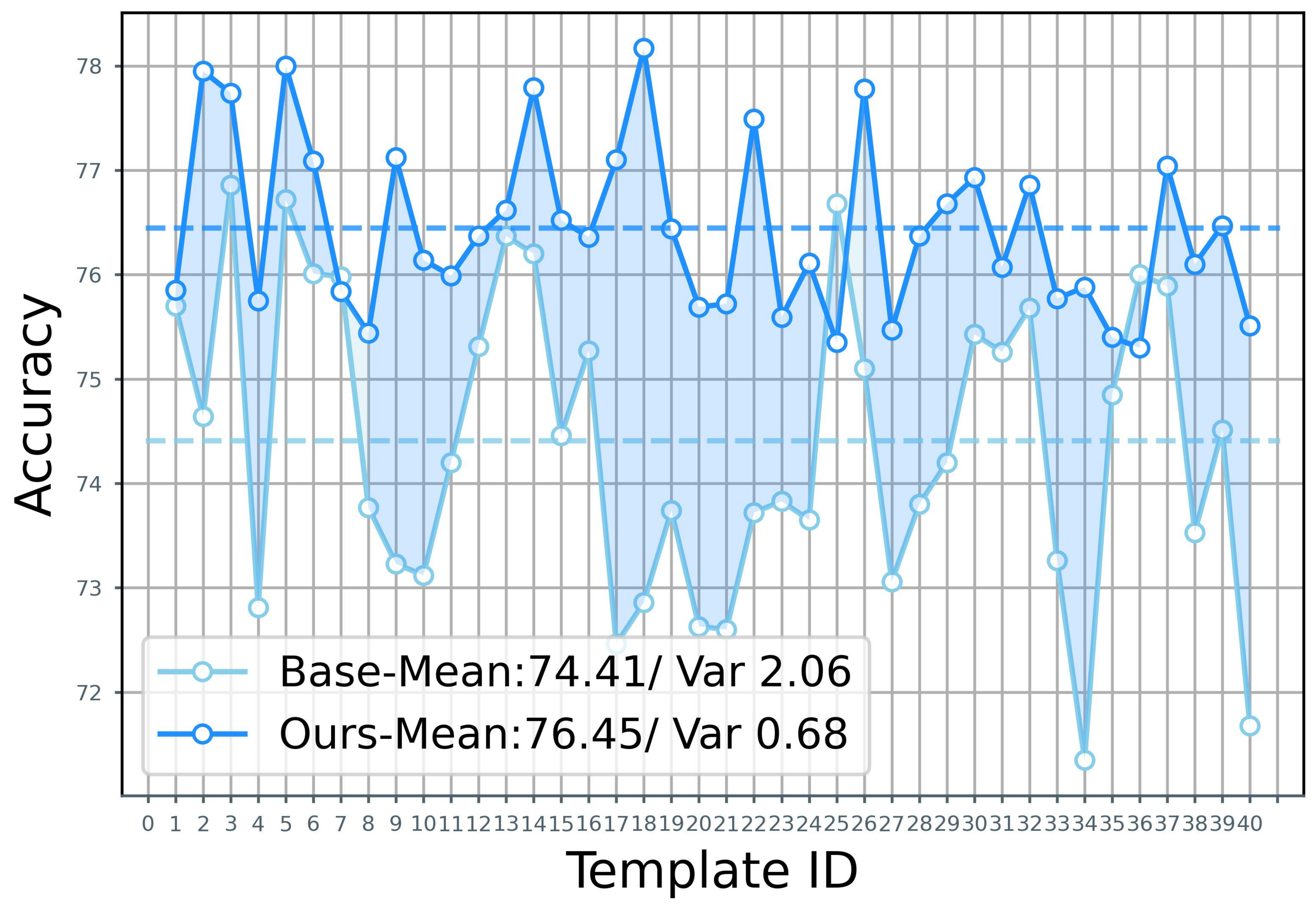}
    }
\end{minipage}
\caption{(Left) Correlation between Template-Sample Similarity and classification accuracy on EuroSAT (correlation coefficients is included in the legend). (Middle) The evolution of the absolute value of correlation coefficients between classification accuracy and template-sample similarity over the course of 1-shot training on EuroSAT. (Right) The effect of different templates on overall performance, shown both before and after applying template correction on EuroSAT.}
\label{fig:intro}
\end{figure*}

\begin{table*}[!h]
\centering
\begin{adjustbox}{max width=0.98\linewidth}
\begin{tabular}{lcccccccccccc}
\toprule
 & ImagNet & SUN & Aircraft & EuroSAT & Cars & Food & Pets & Flowers & Caltech & DTD & UCF & Avg. \\
\midrule
Only class & 64.08 & 60.81 & 20.97 & \textbf{46.38} & 63.67 & 83.94 & 81.92 & 64.67 & 87.22 & \textbf{45.80} & 63.83 & 62.12 \\
w/ Template & \textbf{66.68} & \textbf{62.49} & \textbf{23.25} & 42.38 & \textbf{65.40} & \textbf{85.27} & \textbf{88.36} & \textbf{67.35} & \textbf{93.38} & 44.32 & \textbf{65.15} & 64.00 \\
\midrule
Misclassification Ratio & 4.82 & 6.49 & 4.32 & 9.79 & 5.19 & 2.39 & 1.47 & 4.1 & 1.29 & 8.21 & 6.18 & 4.93 \\
\bottomrule
\end{tabular}
\end{adjustbox} 
\caption{Comparison of classification performance using bare class names versus full templates. We define the \emph{misclassification ratio} as
\textbf{(samples correctly classified using only class names but misclassified after introducing templates) / (total samples)}.
Although templates improve overall accuracy, they also induce errors on a subset of samples that were previously classified correctly.
}
\label{tab:Misclassification}
\end{table*}

\subsection{What is Template-Sample Similarity?}

CLIP-based image classification compares visual features with text embeddings formed by inserting class names into templates. For a given image \(s\), CLIP computes cosine similarities between its visual embedding \(\theta_v(s)\) and three text embeddings: the blank template $t_0 = \texttt{"a photo of a \{\}"}$,
the bare class name \(c\), and the full prompt 
$p_c = \texttt{"a photo of a }c\texttt{"}$.
We denote these metrics as
\begin{equation}
\mathrm{TSS}(s)=\cos\bigl(\theta_t(t_0),\theta_v(s)\bigr),\quad
\end{equation}
\begin{equation}
\mathrm{CSS}(c,s)=\cos\bigl(\theta_t(c),\theta_v(s)\bigr),\quad
\end{equation}
\begin{equation}
\mathrm{PSS}(c,s)=\cos\bigl(\theta_t(p_c),\theta_v(s)\bigr).
\end{equation}
where \(\theta_t\) and \(\theta_v\) are the text and image encoders, respectively. In particular, Template-Sample Similarity (TSS) characterizes the intrinsic compatibility between a visual sample and the general context suggested by the blank template, independent of specific semantic content. Class-Sample Similarity (CSS), on the other hand, directly quantifies the alignment between the visual sample and the semantic embedding of the category name itself. Finally, Prompt-Sample Similarity (PSS) encapsulates both template context and semantic specificity, combining their joint effects into a single metric.

Here, TSS captures how well the image aligns with the generic template context, CSS measures the image’s alignment with the class name itself, and PSS reflects their combined effect. As Figure~\ref{fig:finding}(a) shows, samples with higher TSS consistently achieve higher PSS across classes, indicating a systematic template-driven bias in CLIP’s predictions. By separating TSS (template influence) from CSS (category-specific signal), we can more precisely analyze bias and classification performance in multimodal representation learning.

\begin{figure*}[!tp]
\centering
\includegraphics[width=0.95\linewidth]{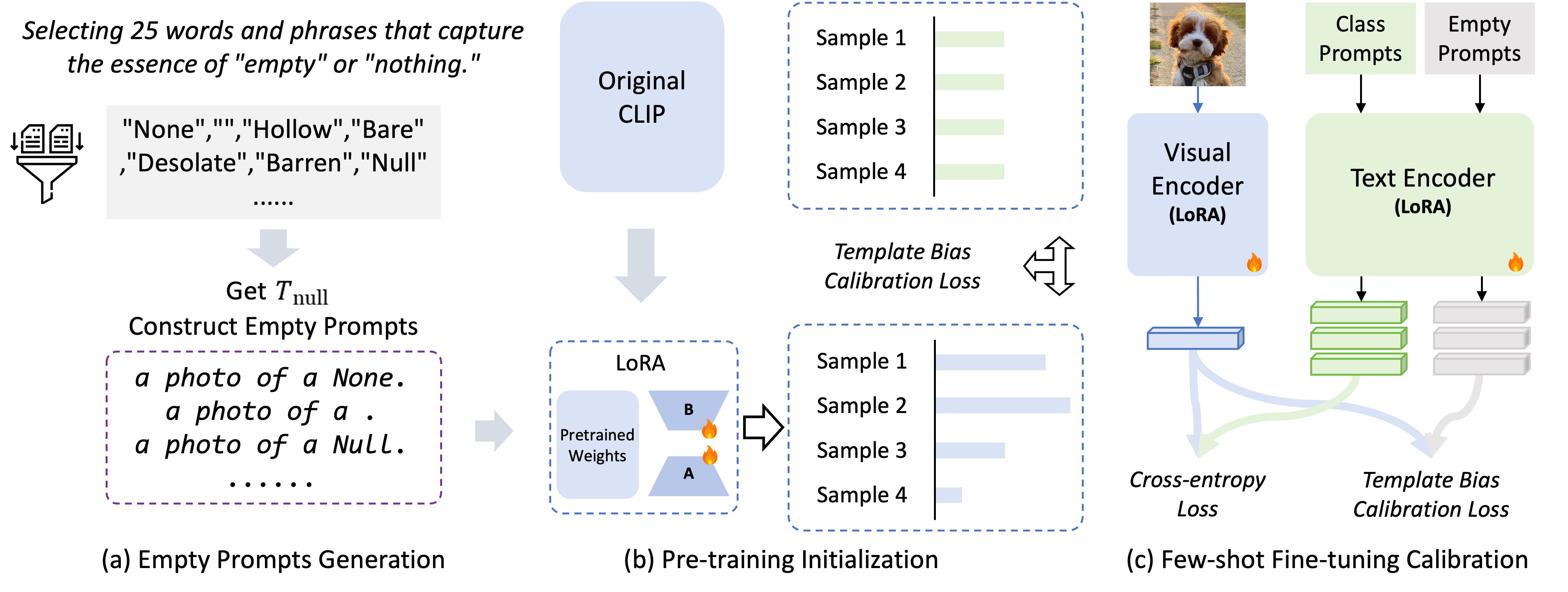}
\caption{The overall framework for template correction involves three main stages. (a) \textbf{Empty Prompts Generation}: A diverse set of empty prompts is manually curated to help identify potential template-induced biases. (b) \textbf{Pretraining Initialization}: These empty prompts are used to detect and correct biases within the CLIP model during the pretraining phase. (c) \textbf{Few-shot Fine-tuning Calibration}: Finally, the model undergoes fine-tuning with few-shot samples to calibrate its performance and improve classification accuracy.} 
\label{fig:our_method}
\end{figure*}

\subsection{Impact of Template-Sample Similarity}

In classification, the essential determinant is the alignment between each visual sample and its true category. We further ask whether the Template-Sample Similarity Score (TSS) also influences the outcome. Table~\ref{tab:Misclassification} shows that appending full templates to the CLIP pipeline consistently raises accuracy across all eleven datasets—boosting average accuracy from 62.12\,\% (using only class names) to 64.00\,\% (with templates). Yet this gain carries a penalty: on average, 4.93\,\% of samples that CLIP correctly classified by class names alone are misclassified when templates are applied. The misclassification ratio peaks at 9.79\,\% on EuroSAT and exceeds 6\,\% for SUN and UCF, while Caltech exhibits the lowest rate at 1.29\,\%. Thus, Table~\ref{tab:Misclassification} highlights a crucial trade-off—\textbf{templates enhance overall accuracy but can also introduce bias.}

To investigate this effect more deeply, we use an unfinetuned CLIP model to compute the TSS for each sample in EuroSAT~\cite{helber2019eurosat} with templates that omit category names. The TSS values are grouped into bins of 400 samples each (e.g., samples with TSS values between 0.19 and 0.20). We then compute the classification accuracy for each bin. Figure~\ref{fig:intro}(Left) shows a strong correlation (correlation coefficient ~0.9) between TSS and the probability of correct classification, indicating that TSS significantly affects whether a sample is accurately classified using a given template.

This phenomenon can be attributed to template-induced bias. As illustrated in Figure~\ref{fig:finding}, when textual features of a template differ significantly from visual characteristics of a sample, the template becomes unsuitable. For example, “a centered satellite photo of {}” aligns well with satellite images' features and help classification, while “a cartoon photo of {}” generates mismatched features and leads to errors. Based on these findings, we draw the following observations.

\begin{observation}
\label{cor:summary1}
A sample’s classification accuracy depends jointly on how well it matches the chosen template and how well it aligns with its true class. In practice, samples that score highly against a template tend to be classified correctly more often when the correct class name is appended. This demonstrates that TSS and CSS act together to determine performance.
\end{observation}

As shown in Figure~\ref{fig:finding}(a), although the TSS captures rich contextual cues, it can inadvertently align with irrelevant visual features during the pre-training alignment phase, causing the model to attend to non-target regions instead of the object of interest. While the precise mechanism by which the template influences performance remains unclear, this observation highlights an important design consideration for few-shot learning algorithms.

\subsection{Can few-shot training eliminate template bias?}

In an ideal classification framework, decision outcomes should depend solely on the degree of alignment between each sample and its true category, remaining invariant to extraneous factors such as TSS. However, as shown in Figure~\ref{fig:intro}(Right), using different templates to construct prompts during training causes substantial fluctuations in model performance, demonstrating that template-induced bias endures despite fine-tuning. To further quantify this effect, we measured the variance in classification accuracy across all categories for each template. Figure~\ref{fig:intro}(Middle) reveals that fine-tuning techniques like LoRA effectively reduce—but do not entirely eliminate—the correlation between classification accuracy and TSS in the CLIP model. Specifically, the correlation coefficient declines after adaptation but stabilizes at approximately 0.2, indicating a residual dependence on template similarity. These findings underscore that direct fine-tuning alone is insufficient to fully remove bias introduced by template prompts.

\begin{observation}
\label{cor:summary2}
Direct fine-tuning alone cannot completely eliminate the bias induced by template prompts, as a non-negligible correlation between classification accuracy and TSS remains.
\end{observation}

\section{Template Correction for Few-Shot Learning}

\subsection{Task Formulation}

In classification tasks using VLMs, we are provided with a set of \( K \) candidate classes. For each class, a textual description, known as a prompt, is created using a predefined template. For example, the template “a photo of a \{\}” can generate prompts like “a photo of a cat” or “a photo of a dog.” This template guides the text encoder to extract features that align with the visual characteristics of each class. Let \( \mathbf{c}_k \) denote the tokenized prompt for the \( k \)-th class. The language encoder \( \theta_t \) processes \( \mathbf{c}_k \) to produce the normalized textual embedding \( \mathbf{t}_k = \theta_t(\mathbf{c}_k) \), which lies on a unit hypersphere. Similarly, each image \( \mathbf{x}_i \) is processed by the visual encoder \( \theta_v \) to obtain the normalized visual embedding \( \mathbf{f}_i = \theta_v(\mathbf{x}_i) \).

Zero-shot prediction is the simplest method for adapting a VLM to downstream tasks, leveraging the model's pretraining \cite{radford2021learning}. For a test image \( \mathbf{x}_i \), the cosine similarity between its visual embedding \( \mathbf{f}_i \) and each class's textual embedding \( \mathbf{t}_k \) is calculated to obtain the prediction logit: $l_{i, k} = \mathbf{f}_i^{\top} \mathbf{t}_k$.
These logits are then transformed into posterior probabilities using the softmax function with temperature scaling \( \tau \):
$p_{i, k} = \exp\left(l_{i, k} / \tau\right) / \sum_{j=1}^K \exp\left(l_{i, j} / \tau\right)$.

The class with the highest posterior probability \( p_{i, k} \) is selected as the predicted label for the image \( \mathbf{x}_i \). This approach allows the VLM to perform classification without requiring additional training, relying solely on the alignment between textual and visual embeddings.

During fine-tuning on downstream tasks, LoRA~\cite{hu2021lora} models the incremental update of pre-trained weights as the product of two small matrices, following the concept of the “intrinsic rank” of a downstream task. In our method, we apply LoRA to both text and visual branches of the CLIP model to fine-tune the original CLIP model, following the settings defined in~\cite{zanella2024low}.

\subsection{Empty Prompts Generation}

To examine CLIP's inherent biases toward specific template structures, we generate empty prompts by combining standard templates with inputs that lack meaningful content. These empty prompts enable us to identify and understand the biases introduced by the templates. The bias information extracted from these prompts is then used to calibrate the CLIP model, thereby enhancing its fairness and accuracy.

Creating a diverse set of meaningless inputs, denoted as \( T_{null} \), is crucial to prevent overfitting to particular instances and to ensure effective calibration. 
First, we select common template structures used in CLIP-based classification tasks, such as “a photo of a \{\}”, then generate 100 words/phrases representing “empty” using GPT-4. For each candidate, we compute the average similarity to all category names and select the top-25 with highest similarity. This ensures selected words lie in the same feature space as category names and remain semantically ambiguous.

The selected 25 words are used to fill these templates, resulting in prompts like “a photo of nothing” or “a photo of emptiness”. We then input these empty prompts into the CLIP model and analyze its responses to detect any inherent biases related to the template structures. Finally, the bias information derived from the empty prompts is used to adjust and calibrate the CLIP model, thereby reducing unwanted template-induced biases.

\subsection{Template Bias Calibration Loss}
As shown in Figure~\ref{fig:finding}, the final classification results are influenced by both the template and the class name because the similarity between the template and the samples varies. To reduce this bias, we introduce a constraint that aligns the similarity relationships between the samples and the template during the classification process.

Figure~\ref{fig:our_method} illustrates our proposed method for removing this bias. We define $\mathcal{M}$ as the current model, and $\mathbf{t}^{E}_{i}$ as the normalized textual embedding for the empty prompt (“a photo of a $T_{null,i}$”). Ideally, the similarity between $\mathbf{t}^{E}_{i}$ and different samples should be consistent. Specifically, if each sample is treated as a separate class in the class set $\mathcal{Y}_x$, using the samples' visual features as classifier weights and $\mathbf{t}^{E}_{i}$ as the query feature, the classification should satisfy the following condition:
\begin{equation}
\mathbb{E}_{\mathbf{t}^{E}_{i}}\left[P\left(y \mid \mathbf{t}^{E}_{i}, \mathcal{ M} ; \forall y \in \mathcal{Y}_x\right)\right] = \frac{1}{|\mathcal{Y}_x|}
\end{equation}
This means that the expected output distribution for $\mathbf{t}^{E}_{i}$ should be uniform across all labels. Let $p^{E}_{i, k} = P\left(y_k \mid \mathbf{t}^{E}_{i}, \mathcal{M} ; \forall y_k \in \mathcal{Y}_x\right)$. To correct for the template bias, we design a loss function $L_{tb}$ based on the above equation:
\begin{equation}
L_{tb} = - \frac{1}{|T_{null}|} \sum_{i=1}^{|T_{null}|} \sum_{k=1}^{|\mathcal{Y}_x|} \frac{1}{|\mathcal{Y}_x|} \ln p^{E}_{i, k}
\end{equation}
$L_{tb}$ ensures that each empty prompt is classified with equal probability for each category associated with the sample. This approach guarantees that the similarity between the instances and the template remains uniform.

\subsection{Training Process}

The training of our model is carried out in two distinct stages: pretraining initialization and few-shot finetuning calibration.

\noindent \textbf{Pretraining Initialization.}
In the first stage, we perform unsupervised pre-training using the loss function \( L_{pre} = L_{tb} \). The main objective of this stage is to initialize the model parameters without any bias. By minimizing \( L_{tb} \), we ensure that the model starts with unbiased initial parameters, which is crucial for preventing template-induced biases from affecting the learning process.

\noindent \textbf{Few-shot Finetuning calibration.}
In the second stage, we fine-tune the model using a few labeled samples. During this phase, we apply the cross-entropy loss function, as commonly used in~\cite{zhou2022learning,zhou2022conditional,zanella2024low}, to adapt the model to specific tasks:
\begin{equation}
L_{ce}= - \frac{1}{N} \sum_{i=1}^N \sum_{k=1}^K y_{ik} \ln p_{i, k}
\end{equation}
Here, \( N \) is the number of samples, \( K \) is the number of classes, \( y_{ik} \) is the ground truth label, and \( p_{i, k} \) is the predicted probability for class \( k \) of sample \( i \).

In addition to \( L_{ce} \), we retain the template bias calibration loss \( L_{tb} \) during fine-tuning to maintain the model's unbiasedness with respect to the template. The total loss function for fine-tuning is therefore defined as:
\begin{equation}
L_{fine} = L_{ce} + \alpha L_{tb}
\end{equation}
where \( \alpha \) is a hyperparameter that controls the weight of the bias calibration loss relative to the cross-entropy loss.

This two-stage training process ensures that the model not only learns to perform accurately on the target task with limited labeled data but also remains free from biases introduced by the template structures. By first establishing unbiased initial parameters and then carefully fine-tuning with a balanced loss function, our approach achieves both fairness and high performance in classification tasks. Please refer to the Appendix for the pseudocode of the algorithm.

        
        
        
        
        
        

\section{Experiments}

\subsection{Experimental Setup}

\begin{table*}[!htbp]
\centering
\begin{adjustbox}{max width=0.95\linewidth}
\begin{tabular}{*{15}{c}}  
\toprule 
 \multirow{1}*{Shots} & \multirow{1}*{Method}  &  \multirow{1}*{ImageNet}  & \multirow{1}*{SUN}
 & \multirow{1}*{Aircraft}  & \multirow{1}*{EuroSAT}  & \multirow{1}*{Cars}  
  & \multirow{1}*{Food}   & \multirow{1}*{Pets}  & \multirow{1}*{Flowers}  
  & \multirow{1}*{Caltech}  & \multirow{1}*{DTD} & \multirow{1}*{UCF} & \multirow{1}*{Average}\\
\midrule
& CoOp(4) & 68.0 & 67.3 & 26.2 & 50.9  & 67.1 & 82.6  & 90.3 & 72.7 & 93.2 & 50.1 & 70.7 & 67.2\\
& CoOp(16) & 65.7 & 67.0 & 20.8 & 56.4  & 67.5 & 84.3  & 90.2 & 78.3 & 92.5 & 50.1 & 71.2 & 67.6\\
& CoCoOp & 69.4 & 68.7 & 28.1 & 55.4  & 67.6 & 84.9  & 91.9 & 73.4 & 94.1 & 52.6 & 70.4 & 68.8\\
& TIP-Adapter-F & 69.4 & 67.2 & 28.8 & 67.8 & 67.1 & 85.8 & 90.6 & \textbf{83.8} & 94.0 & 51.6 & 73.4 & 70.9\\
& CLIP-Adapter & 67.9 & 65.4 & 25.2 & 49.3 & 65.7 & 86.1 & 89.0 & 71.3 & 92.0 & 44.2 & 66.9 & 65.7\\
\textbf{1} & PLOT++ & 66.5 & 66.8 & 28.6 & 65.4 & 68.8 & 86.2 & 91.9 & 80.5 & 94.3 & 54.6 & 74.3 & 70.7\\
& KgCoOp & 68.9 & 68.4 & 26.8 & 61.9 & 66.7 & \textbf{86.4} & 92.1 & 74.7 & 94.2 & 52.7 & 72.8 & 69.6\\
& TaskRes & 69.6 & 68.1 & 31.3 & 65.4 & 68.8 & 84.6 & 90.2 & 81.7 & 93.6 & 53.8 & 71.7 & 70.8\\
& MaPLe & 69.7 & 69.3 & 28.1 & 29.1 & 67.6 & 85.4 & 91.4 & 74.9 & 93.6 & 50.0 & 71.1 & 66.4\\
& ProGrad & 67.0 & 67.0 & 28.8 & 57.0 & 68.2 & 84.9 & 91.4 & 80.9 & 93.5 & 52.8 & 73.3 & 69.5\\
& LoRA-CLIP & 70.4 & 70.4 & 30.2 & 72.3  & 70.1 & 84.3  & 92.3 & 83.2 & 93.7 & 54.3 & \textbf{76.3} & 72.5\\
& \textbf{OURS} & \textbf{70.5} & \textbf{70.5} & \textbf{32.2} & \textbf{78.4}  & \textbf{70.6} & 85.7  & \textbf{92.9} & 83.3 & \textbf{94.8} & \textbf{54.5} & 76.0 & \textbf{73.6}\\
\bottomrule

&CoOp(4) &68.7&68.0&28.1&66.2&70.5&82.6&89.9&80.9&93.0&53.7&73.5&70.5\\
&CoOp(16) &67.0&67.0&25.9&65.1&70.4&84.4&89.9&88.0&93.1&54.1&74.1&70.8\\
&CoCoOp &70.1&69.4&29.3&61.8&68.4&85.9&91.9&77.8&94.4&52.3&73.4&70.4\\
&TIP-Adapter-F &70.0&68.6&32.8&73.2&70.8&86.0&91.6&90.1&93.9&57.8&76.2&73.7\\
&CLIP-Adapter &68.2&67.2&27.0&51.2&66.6&86.2&89.7&71.7&93.4&45.4&68.4&66.8\\
\textbf{2} &PLOT++ &68.3&68.1&31.1&76.8&73.2&86.3&92.3&89.8&94.7&56.7&76.8&74.0\\
&KgCoOp &69.6&69.6&28.0&69.2&68.2&\textbf{86.6}&\textbf{92.3}&79.8&94.5&55.3&74.6&71.6\\
&TaskRes &70.2&70.5&32.7&70.2&72.1&85.6&90.7&84.4&94.3&55.6&75.2&72.9\\
&MaPLe &70.0&70.7&29.5&59.4&68.5&86.5&91.8&79.8&94.9&50.6&74.0&70.5\\
&ProGrad &69.1&69.0&31.1&66.3&72.4&84.8&91.5&87.5&93.6&56.0&75.6&72.4\\
&CLIP-LoRA &70.8&71.3&33.2&82.7&73.2&83.2&91.3&89.8&94.6&59.9&\textbf{80.0}&75.5\\
& Ours & \textbf{71.0} & \textbf{71.7} & \textbf{34.9}	& \textbf{83.3}	& \textbf{73.5}	& 86.2	& 91.9	& \textbf{90.4}	& \textbf{95.1}	& \textbf{61.4} & 79.8	& \textbf{76.3} \\

\bottomrule
&CoOp (4) & 69.7 & 70.6 & 29.7 & 65.8 & 73.4 & 83.5 & 92.3 & 86.6 & 94.5 & 58.5 & 78.1 & 73.0\\
&CoOp (16)  & 68.8 & 69.7 & 30.9 & 69.7 & 74.4 & 84.5 & 92.5 & 92.2 & 94.5 & 59.5 & 77.6 & 74.0\\
&CoCoOp  & 70.6 & 70.4 & 30.6 & 61.7 & 69.5 & 86.3 & 92.7 & 81.5 & 94.8 & 55.7 & 75.3 & 71.7\\
&TIP-Adapter-F  & 70.7 & 70.8 & 35.7 & 76.8 & 74.1 & 86.5 & 91.9 & 92.1 & 94.8 & 59.8 & 78.1 & 75.6\\
&CLIP-Adapter  & 68.6 & 68.0 & 27.9 & 51.2 & 67.5 & 86.5 & 90.8 & 73.1 & 94.0 & 46.1 & 70.6 & 67.7\\
\textbf{4} &PLOT++  & 70.4 & 71.7 & 35.3 & 83.2 & 76.3 & 86.5 & 92.6 & 92.9 & 95.1 & 62.4 & 79.8 & 76.9\\
&KgCoOp  & 69.9 & 71.5 & 32.2 & 71.8 & 69.5 & 86.9 & 92.6 & 87.0 & 95.0 & 58.7 & 77.6 & 73.9\\
&TaskRes  & 71.0 & 72.7 & 33.4 & 74.2 & 76.0 & 86.0 & 91.9 & 85.0 & 95.0 & 60.1 & 76.2 & 74.7\\
&MaPLe  & 70.6 & 71.4 & 30.1 & 69.9 & 70.1 & 86.7 & 93.3 & 84.9 & 95.0 & 59.0 & 77.1 & 73.5\\
&ProGrad & 70.2 & 71.7 & 34.1 & 69.6 & 75.0 & 85.4 & 92.1 & 91.1 & 94.4 & 59.7 & 77.9 & 74.7\\
&LoRA-CLIP & 71.4 & 72.8 & 37.9 & 84.9  & 77.4 & 82.7  & 91.0 & 93.7 & 95.2 & 63.8 & 81.1 & 77.4\\

&\textbf{OURS} & \textbf{71.5} & \textbf{74.0} & \textbf{40.2} & \textbf{87.6}  & \textbf{77.7} & \textbf{87.1}  & \textbf{93.7 }& \textbf{94.8} & \textbf{95.7} & \textbf{65.6} & \textbf{82.2} & \textbf{79.1}\\
\bottomrule
&CoOp(4) & 70.8 & 72.4 & 37.0 & 74.7 & 76.8 & 83.3 & 92.1 & 95.0 & 94.7 & 63.7 & 79.8 & 76.4\\
 &CoOp(16) & 70.6 & 71.9 & 38.5 & 77.1 & 79.0 & 82.7 & 91.3 & 94.9 & 94.5 & 64.8 & 80.0 & 76.8\\
 &CoCoOp & 70.8 & 71.5 & 32.4 & 69.1 & 70.4 & 87.0 & 93.3 & 86.3 & 94.9 & 60.1 & 75.9 & 73.8\\
 &TIP-Adapter-F & 71.7 & 73.5 & 39.5 & 81.3 & 78.3 & 86.9 & 91.8 & 94.3 & 95.2 & 66.7 & 82.0 & 78.3\\
 &CLIP-Adapter & 69.1 & 71.7 & 30.5 & 61.6 & 70.7 & 86.9 & 91.9 & 83.3 & 94.5 & 50.5 & 76.2 & 71.5\\
 \textbf{8} &PLOT++ & 71.3 & 73.9 & 41.4 & 88.4 & 81.3 & 86.6 & 93.0 & 95.4 & 95.5 & 66.5 & 82.8 & 79.6\\
 &KgCoOp & 70.2 & 72.6 & 34.8 & 73.9 & 72.8 & 87.0 & 93.0 & 91.5 & 95.1 & 65.6 & 80.0 & 76.0\\
 &TaskRes & 72.3 & 74.6 & 40.3 & 77.5 & 79.6 & 86.4 & 92.0 & 96.0 & 95.3 & 66.7 & 81.6 & 78.4\\
 &MaPLe & 71.3 & 73.2 & 33.8 & 82.8 & 71.3 & 87.2 & 93.1 & 90.5 & 95.1 & 63.0 & 79.5 & 76.4\\
 &ProGrad & 71.3 & 73.0 & 37.7 & 77.8 & 78.7 & 86.1 & 92.2 & 95.0 & 94.8 & 63.9 & 80.5 & 77.4\\
 &CLIP-LoRA & 72.3 & 74.7 & 45.7 & 89.7 & 82.1 & 83.1 & 91.7 & 96.3 & 95.6 & 67.5 & 84.1 & 80.3\\
 
&\textbf{OURS } & \textbf{72.3} & \textbf{75.5} & \textbf{49.7} & \textbf{91.2} & \textbf{82.5} & \textbf{87.3} & \textbf{94.2} & \textbf{96.9} & \textbf{95.9} & \textbf{69.2} & \textbf{85.0} & \textbf{81.8}\\
\bottomrule
&CoOp(4)&71.5&74.6&40.1&83.5&79.1&85.1&92.4&96.4&95.5&69.2&81.9&79.0\\
&CoOp(16)&71.9&74.9&43.2&85.0&82.9&84.2&92.0&96.8&95.8&69.7&83.1&80.0\\
&CoCoOp&71.1&72.6&33.3&73.6&72.3&87.4&93.4&89.1&95.1&63.7&77.2&75.4\\
&TIP-Adapter-F&73.4&76.0&44.6&85.9&82.3&86.8&92.6&96.2&95.7&70.8&83.9&80.7\\
&CLIP-Adapter&69.8&74.2&34.2&71.4&74.0&87.1&92.3&92.9&94.9&59.4&80.2&75.5\\
\textbf{16} &PLOT++&72.6&76.0&46.7&92.0&84.6&87.1&93.6&97.6&96.0&71.4&85.3&82.1\\
&KgCoOp&70.4&73.3&36.5&76.2&74.8&87.2&93.2&93.4&95.2&68.7&81.7&77.3\\
&TaskRes&73.0&76.1&44.9&82.7&83.5&86.9&92.4&97.5&95.8&71.5&84.0&80.8\\
&MaPLe&71.9&74.5&36.8&87.5&74.3&87.4&93.2&94.2&95.4&68.4&81.4&78.6\\
&ProGrad&72.1&75.1&43.0&83.6&82.9&85.8&92.8&96.6&95.9&68.8&82.7&79.9\\
&CLIP-LoRA&73.6&76.1&54.7&92.1&86.3&84.2&92.4&98.0&96.4&72.0&\textbf{86.7}&83.0\\
& Ours & \textbf{73.6} & \textbf{76.4} & \textbf{57.4}	& \textbf{92.4}	& \textbf{86.3}	& \textbf{87.6} & \textbf{94.6}	& \textbf{98.2}	& \textbf{96.5}	& \textbf{72.8} & 86.5	& \textbf{83.8} \\

\bottomrule
\end{tabular}
\end{adjustbox}

\caption{
Detailed results for 11 datasets with the ViT-B/16 as visual backbone.Top-1 accuracy averaged over 3 random seeds is reported. Highest value is high lighted inbold.
}
\label{tab:fewshot_1shot}
\end{table*}

\noindent \textbf{Dataset.}
Following the setup of previous work~\cite{zhou2022learning,zanella2024low}, we evaluate our approach on 11 datasets covering various fine-grained classification tasks: scenes (SUN397~\cite{xiao2010sun}), aircraft types (Aircraft~\cite{maji2013fine}), satellite imagery (EuroSAT~\cite{helber2019eurosat}), automobiles (StanfordCars~\cite{krause20133d}), food items (Food101~\cite{bossard2014food}), pet breeds (OxfordPets~\cite{parkhi2012cats}), flowers (Flower102~\cite{nilsback2008automated}), general objects (Caltech101~\cite{fei2004learning}), textures (DTD~\cite{cimpoi2014describing}), and human actions (UCF101~\cite{soomro2012ucf101}), in addition to ImageNet~\cite{deng2009imagenet}. These datasets provide a comprehensive benchmarking framework for evaluating few-shot visual classification tasks.

\noindent \textbf{Comparative Methods.}
We compare our model against several prompt-based methods, including CoOp~\cite{zhou2022learning} with 4 learnable tokens, CoOp~\cite{zhou2022learning} with 16 learnable tokens, CoCoOp~\cite{zhou2022conditional}, PLOT++~\cite{chen2022plot} (an adaptation of the original PLOT designed for transformer architectures), KgCoOp~\cite{yao2023visual}, MaPLe~\cite{khattak2023maple}, and ProGrad~\cite{zhu2023prompt} with 16 tokens. Additionally, we evaluate adapter-based methods such as Tip-Adapter-F~\cite{zhang2022tip} and TaskRes~\cite{yu2023task}, as well as the LoRA-based method CLIP-LoRA~\cite{zanella2024low}. These comparative methods provide a comprehensive benchmark to assess the effectiveness of our proposed approach in few-shot visual classification tasks.


\subsection{Few-shot Learning Result}

Table~\ref{tab:fewshot_1shot} presents the few-shot learning performance of various models across 11 datasets. Among the baselines, LoRA is the most competitive, achieving the second-best overall performance. Our model, which adds template bias correction to LoRA, shows superior or comparable results on most datasets, with significant improvements on EuroSAT, Food101, OxfordPet, DTD, and Aircraft. On average, our model improves performance by approximately 1.5 points over LoRA across the 11 datasets.
\begin{table}[!t]
    \centering
    \begin{adjustbox}{max width=0.95\linewidth}
    \begin{tabular}{ccccccc}
    \toprule
    Pretrain & - & - & - & $\checkmark$ & $\checkmark$ & $\checkmark$ \\
    Multiple & - & - & $\checkmark$ & - & $\checkmark$ & $\checkmark$ \\
    $+L_{tb}$ & - & $\checkmark$ & $\checkmark$ & - & - & $\checkmark$ \\
    \midrule
    1-shot & 72.3 & 73.9 & 76.2 & 73.3 & 74.2 & 78.4 \\
    4-shot & 84.9 & 85.8 & 86.9 & 86.4 & 87.9 & 87.6 \\
    \bottomrule
    \end{tabular}
    \end{adjustbox} 
\caption{Ablating main components of our model on EuroSAT.}
\label{tab:ablation} 
\end{table}

\begin{table}[!t]  
    \centering
    \begin{adjustbox}{max width=0.98\linewidth}
    \begin{tabular}{*{7}{c}}  
    \toprule 
      \multirow{1}*{Method}  &  \multirow{1}*{ImageNet}  & \multirow{1}*{SUN}
     & \multirow{1}*{Aircraft}  & \multirow{1}*{EuroSAT}   & \multirow{1}*{DTD}  & \multirow{1}*{Average}\\
     \midrule
      Baseline & 70.4 & 70.4 & 30.2 & 72.3 & 54.3 & 59.5 \\
      Pull Closer & 69.5 & 69.3 & 30.0 & 71.6 & 51.9 & 58.5 \\
      Push Away & 68.4 & 65.8 & 7.9 & 62.9 & 48.4 & 50.7 \\
      Ours & 70.5 & 70.5 & 32.2 & 78.3 & 54.5 & 61.2 \\
    \bottomrule
    \end{tabular}
    \end{adjustbox} 
\caption{Evaluation of three TSS modification strategies.}
\label{tab:alternative}
\end{table}


\subsection{Ablation Study}
We conduct ablation study to evaluate the contribution of each module, as shown in Table~\ref{tab:ablation}. The modules are defined as follows: \textbf{Pretrain}: The pretraining initialization phase; \textbf{Multiple}: The use of multiple empty prompts to create a diverse set of prompts; \bm{$+L_{tb}$}: The incorporation of the $L_{tb}$ loss during the few-shot fine-tuning phase. Our results demonstrate that each module enhances performance.


In our method, we eliminate the bias introduced by the template by ensuring a consistent distance between each sample and the “empty template” (i.e., enforcing a consistent TSS). To demonstrate the effectiveness of this design, we further compared two alternative strategies for modifying TSS: \textbf{Pull Closer}, which reduces the distance between each sample and the “empty template” (increasing TSS), and\textbf{ Push Away}, which increases the distance (decreasing TSS). The results, shown in Table~\ref{tab:alternative}, indicate that neither the Pull Closer nor the Push Away strategies lead to performance improvements. In contrast, our method effectively enhances the model's performance.

\subsection{Mitigating Template-Induced Bias}
As discussed in the introduction, template-induced bias can significantly impact the performance of few-shot learning models. In this section, we provide a detailed analysis to evaluate how effectively our approach mitigates this bias. 
\begin{figure}[!t]
\centering
\includegraphics[width=0.90\linewidth]{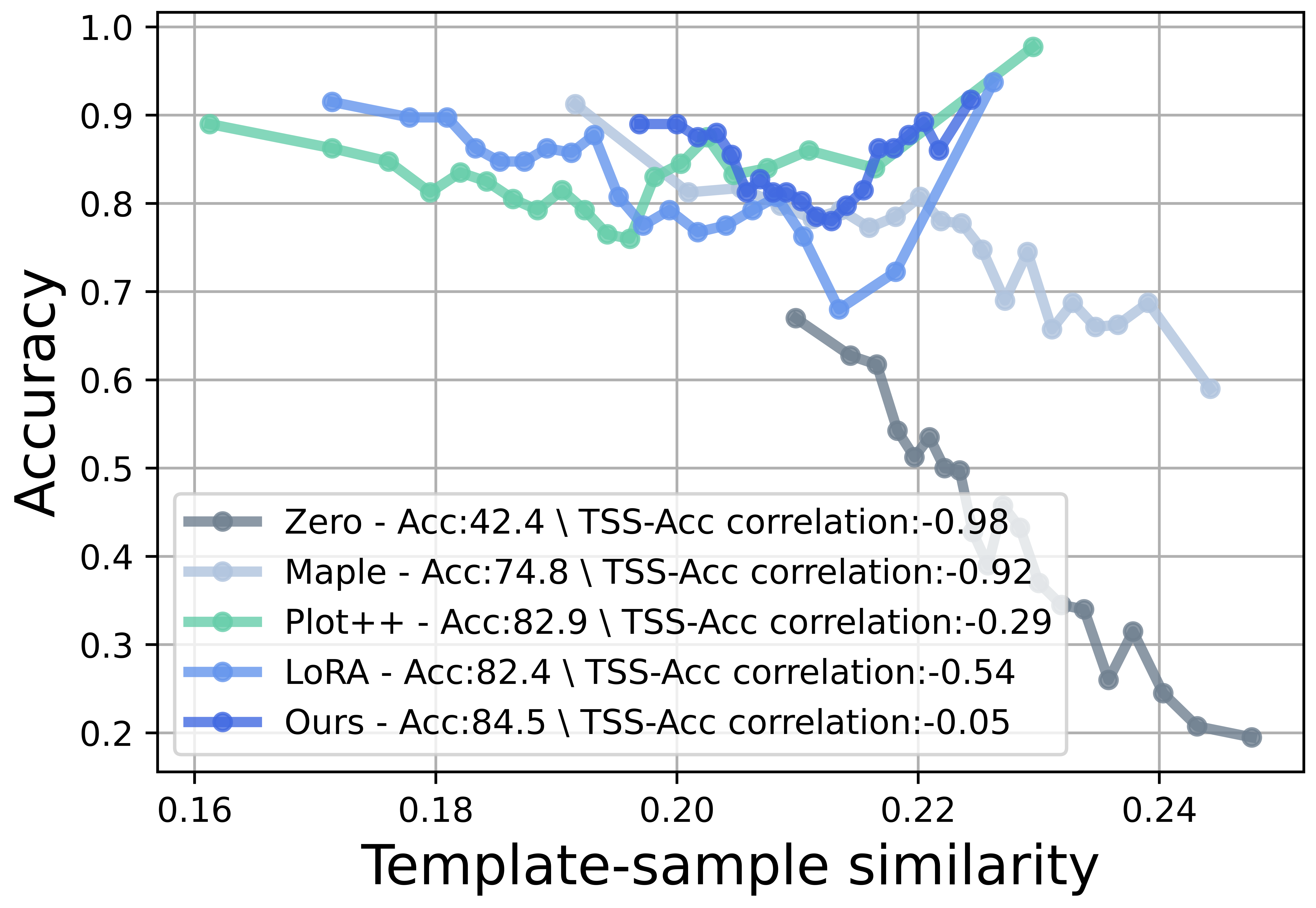}
\caption{The relationship between template-sample similarity and classification accuracy under different training methods. Result for 4-shot finetuning on EuroSAT, seed 1.} 
\label{fig:zero_lora_ours_TSs-Acc_result}
\end{figure}
As shown in Figure~\ref{fig:zero_lora_ours_TSs-Acc_result}, the zero-shot CLIP model exhibits a strong negative correlation (-0.98) between TSS and classification accuracy. Training with LoRA reduces this correlation to -0.54 while improving performance. In contrast, our method nearly eliminates this correlation (-0.05) and achieves the highest classification accuracy among the compared models.
Figure~\ref{fig:zero_lora_ours_TSs-Acc_result} shows that the zero-shot CLIP model has a strong negative correlation (-0.98) between TSS and accuracy. LoRA training reduces this correlation to -0.54 and improves performance. Our method almost eliminates this correlation (-0.05) and achieves the highest accuracy. Figure~\ref{fig:intro} (Middle) shows our method provides a lower-bias initialization, with a TSS-accuracy correlation of 0.38 at iteration 0, compared to the original model's 0.85. Throughout training, our model maintains a lower correlation, showing that the constraint loss reduces template bias and focuses on sample-specific information, leading to a 2-point improvement over LoRA-only training. Figure~\ref{fig:intro} (Right) illustrates the impact of templates on performance. LoRA-trained models are affected by template bias, while our model remains robust to template variations and consistently outperforms LoRA-trained models. Our method effectively mitigates template-induced bias, improving performance.

\section{Related Work}
\label{sec:related}
VLMs combine visual and textual data, making bias evaluation particularly challenging. Models like CLIP~\cite{radford2021learning} and LLaVA~\cite{liu2024visual} can inherit biases from the images and texts used during training. These biases arise from both visual and textual sources, affecting how the model behaves. Several studies~\cite{fabbrizzi2022survey,cabello2023evaluating,seth2023dear,zhang2024micm} have examined biases in VLMs caused by imbalanced datasets and stereotypes, leading to biased representations of gender, race, and professions. To evaluate these biases, datasets such as PAIRS~\cite{fraser2024examining}, which includes images of individuals from diverse racial and gender backgrounds, and VisoGender~\cite{hall2024visogender}, which looks at gender bias in pronoun resolution, have been created. Other research uses specially designed inputs to detect biases in VLMs. For instance, \cite{leng2024mitigating} investigates bias in the text branch by introducing noisy samples to the visual branch. Similarly, \cite{zhang2024debiasing} examines bias in the text branch by providing empty visual inputs. Additionally, \cite{chuang2023debiasing} uses prompts containing biased terms to identify biased directions in text embeddings. While these studies focus on the inherent biases within VLMs, few have explored biases that may arise when VLMs are applied to downstream tasks~\cite{zhang2024learning,zhang2024micm,zou2022margin,zou2024closer,zou2024flatten,zou2021annotation,xiaoevery,zeng2025objects,chen2025automated,zhang2024learning}, such as those introduced by templates. To our knowledge, this study is the first to identify and investigate the biases caused by the template.

\section{Conclusion}
\label{sec:con}
In conclusion, we reveal that TSS introduces significant bias in CLIP. To address this issue, we propose a novel method that uses "empty" prompts to correct the bias introduced by TSS. Extensive experiments demonstrate the effectiveness of our method, highlighting the importance of addressing template-induced biases for improved few-shot learning.

\section{Acknowledgments}
This work is supported by the National Key Research and Development Program of China under grant 2024YFC3307900; the National Natural Science Foundation of China under grants 62436003, 62206102, 62376103, 62302184, and 62402015; the Major Science and Technology Project of Hubei Province under grant 2025BAB011 and 2024BAA008; the Hubei Science and Technology Talent Service Project under grant 2024DJC078; Ant Group through the CCF–Ant Research Fund; the Postdoctoral Fellowship Program of the China Postdoctoral Science Foundation under grant GZB20230024; and the China Postdoctoral Science Foundation under grant 2024M750100. Computations were performed on the HPC Platform of Huazhong University of Science and Technology.

\bibliography{aaai2026}
\clearpage

\appendix
\title{Decoupling Template Bias in CLIP: Harnessing Empty Prompts for Enhanced Few-Shot Learning}
\section{Empty Prompt}
We employ the following 25 terms to represent “empty”: “None", “ ”, “Vacant”, “BlankVoid”, “Hollow”, “Bare”, “Desolate”, “Barren”, “Null”, “Naked”, “Devoid”, “Vacuous”, “Unoccupied”, “Sparse”, “Clean”, “Clear”, “Abandoned”, “Forsaken”, “Deserted”, “Uninhabited”, “Unused”, “Vast”, “Sterile”, “Unfilled”, “Unpopulated”. For example, using the template “a photo of a {}”, the corresponding empty prompts are:

\lstset{
    basicstyle=\ttfamily\color{cyan}, 
    keywordstyle=\color{cyan},         
    commentstyle=\color{cyan},         
    stringstyle=\color{cyan},          
    numberstyle=\tiny\color{cyan},     
    frame=single,                      
    breaklines=true,
    linewidth=0.99\linewidth   
}

\lstset{language=python} 
\begin{lstlisting}
a photo of a None.
a photo of a .
a photo of a Vacant.
a photo of a BlankVoid.
a photo of a Hollow.
a photo of a Bare.
a photo of a Desolate.
a photo of a Barren.
a photo of a Null.
a photo of a Naked.
a photo of a Devoid.
a photo of a Vacuous.
a photo of a Unoccupied.
a photo of a Sparse.
a photo of a Clean.
a photo of a Clear.
a photo of a Abandoned.
a photo of a Forsaken.
a photo of a Deserted.
a photo of a Uninhabited.
a photo of a Unused.
a photo of a Vast.
a photo of a Sterile.
a photo of a Unfilled.
a photo of a Unpopulated.
\end{lstlisting}

\section{More Analysis}
\subsection{Multiple Template}
Our work emphasizes that template-induced biases can suppress model performance. It is crucial to clarify that our study does not discount the value of templates. Templates provide task-specific contextual information, which can significantly enhance performance in downstream tasks. However, our focus is on addressing the biases embedded within the prior knowledge imparted by templates, which can hinder model effectiveness. These biases are intrinsic to the use of templates and cannot be mitigated merely by choosing different templates.

Figure 2(Right) in the main text illustrates the performance variations of both the LoRA model~\cite{zanella2024low} and our proposed model across different template selections. The results show that the LoRA model's performance is highly sensitive to the choice of template, exhibiting substantial variability across templates. In contrast, our model consistently outperforms the LoRA model for nearly all templates, demonstrating that our debiasing mechanism effectively mitigates template-induced biases irrespective of template selection. This highlights that such biases persist regardless of the specific template used.

To further substantiate our findings, we employed seven widely used templates~\cite{radford2021learning} to construct prompts for each category. By averaging their features, we derived a text representation for each category, thereby eliminating biases associated with individual template selection and ensuring robust contextual information across datasets. The fine-tuning results, presented in Table~\ref{tab:fewshot_7template}, reveal that while multiple templates do not always outperform a single template across all datasets, our model consistently achieves performance gains in both single-template and multi-template settings.

It is worth noting that Table 1 excludes results for the ImageNet~\cite{deng2009imagenet}, Stanford Cars~\cite{krause20133d}, and SUN397~\cite{xiao2010sun} datasets due to computational constraints. Specifically, generating prompts using seven templates per category for these datasets exceeds the 24GB memory limit, making it infeasible to conduct experiments with multiple templates for these large-scale datasets.

\begin{table*}[h]
\caption{
Detailed results for 8 datasets with the ViT-B/16 as visual backbone.Top-1 accuracy averaged over 3 random seeds is reported. Highest value is high lighted inbold. \textbf{(7 template)} means that we use 7 templates at the same time to build 7 prompts for each category, and take the mean of their features as the final text feature for this category.
}
\label{tab:fewshot_7template}    
\centering
\begin{adjustbox}{max width=0.95\linewidth}
\begin{tabular}{*{12}{c}}  
\toprule 
 \multirow{1}*{Shots} & \multirow{1}*{Method} 
 & \multirow{1}*{Aircraft}  & \multirow{1}*{EuroSAT}
  & \multirow{1}*{Food}   & \multirow{1}*{Pets}  & \multirow{1}*{Flowers}  
  & \multirow{1}*{Caltech}  & \multirow{1}*{DTD} & \multirow{1}*{UCF} & \multirow{1}*{Average}\\
\midrule
& LoRA-CLIP & 30.2 & 72.3  & 84.3  & 92.3 & 83.2 & 93.7 & 54.3 & \textbf{76.3} & 73.3\\
1 & \textbf{OURS}  & \textbf{32.2} & \textbf{78.4}  & 85.7  & \textbf{92.9} & 83.3 & \textbf{94.8} & 54.5 & 76.0 & 74.7\\
& LoRA-CLIP (7 template) & 29.9 & 72.9 & 84.8  & 91.8 & \textbf{83.7} & 94.1 & 55.0 & 75.9 & 73.5\\
& \textbf{OURS (7 template)}  & 31.5 & 78.1  & \textbf{86.1}  & 92.6 & 82.8 & 94.5 & \textbf{55.6} & 76.2 & \textbf{74.8}\\
\bottomrule
&LoRA-CLIP  & 37.9 & 84.9   & 82.7  & 91.0 & 93.7 & 95.2 & 63.8 & 81.1 & 78.8\\
4 &\textbf{OURS}  & \textbf{40.2} & \textbf{87.6}   & 87.1  & \textbf{93.7} & 94.8 & \textbf{95.7} & 65.6 & 82.2 & \textbf{80.9}\\
& LoRA-CLIP (7 template) & 36.8 & 86.1   & 82.9  & 90.3 & 93.9 & 95.1 & 64.0 & 80.3 & 78.7\\
&\textbf{OURS (7 template)}  & 39.7 & 87.2   & \textbf{87.6}  & 93.5 & \textbf{94.9} & 95.5 & \textbf{65.8} & \textbf{82.4} & 80.8\\

\bottomrule
&CLIP-LoRA  & 45.7 & 89.7  & 83.1 & 91.7 & 96.3 & 95.6 & 67.5 & 84.1 & 81.7\\
8 &\textbf{OURS } & \textbf{49.4} & \textbf{91.2} & 87.3 & 94.2 & \textbf{96.9} & 95.9 & 69.2 & \textbf{85.0} & \textbf{83.6}\\
&CLIP-LoRA (7 template)  & 46.2 & 90.1  & 83.1 & 91.1 & 95.8 & 95.8 & 68.1 & 83.2 & 81.6\\
&\textbf{OURS (7 template)} & 48.6 & 90.8 & \textbf{87.8} & \textbf{94.7} & 96.2 & \textbf{96.0} & \textbf{69.7} & 84.2 & 83.5\\
\bottomrule
\end{tabular}
\end{adjustbox}
\end{table*}

\subsection{Hyperparameter Study}
\begin{figure}[!htp]
\centering
\includegraphics[width=1.0\linewidth]{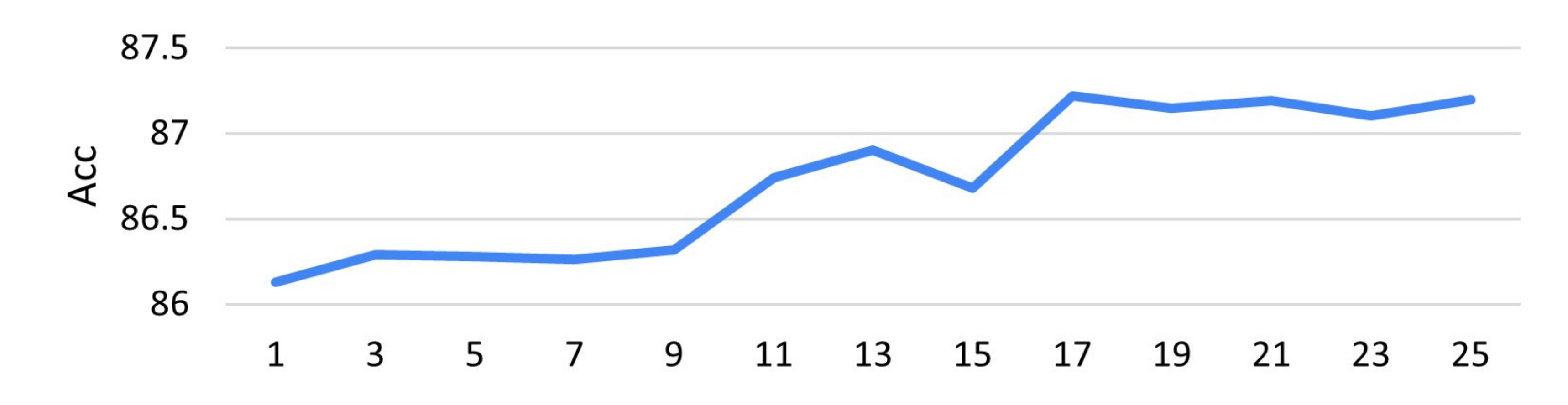}
\caption{Impact of empty prompt count on performance on EuroSAT.} 
\label{fig:empty}
\end{figure}

Our method introduces a single hyperparameter, $N$, which denotes the number of empty prompts utilized. As illustrated in Figure~\ref{fig:empty}, the performance of the model is significantly influenced by variations in $N$. Specifically, increasing the number of empty prompts reduces performance fluctuations, resulting in more stable and consistent improvements.

\begin{figure}[!t]
\centering
\includegraphics[width=\linewidth]{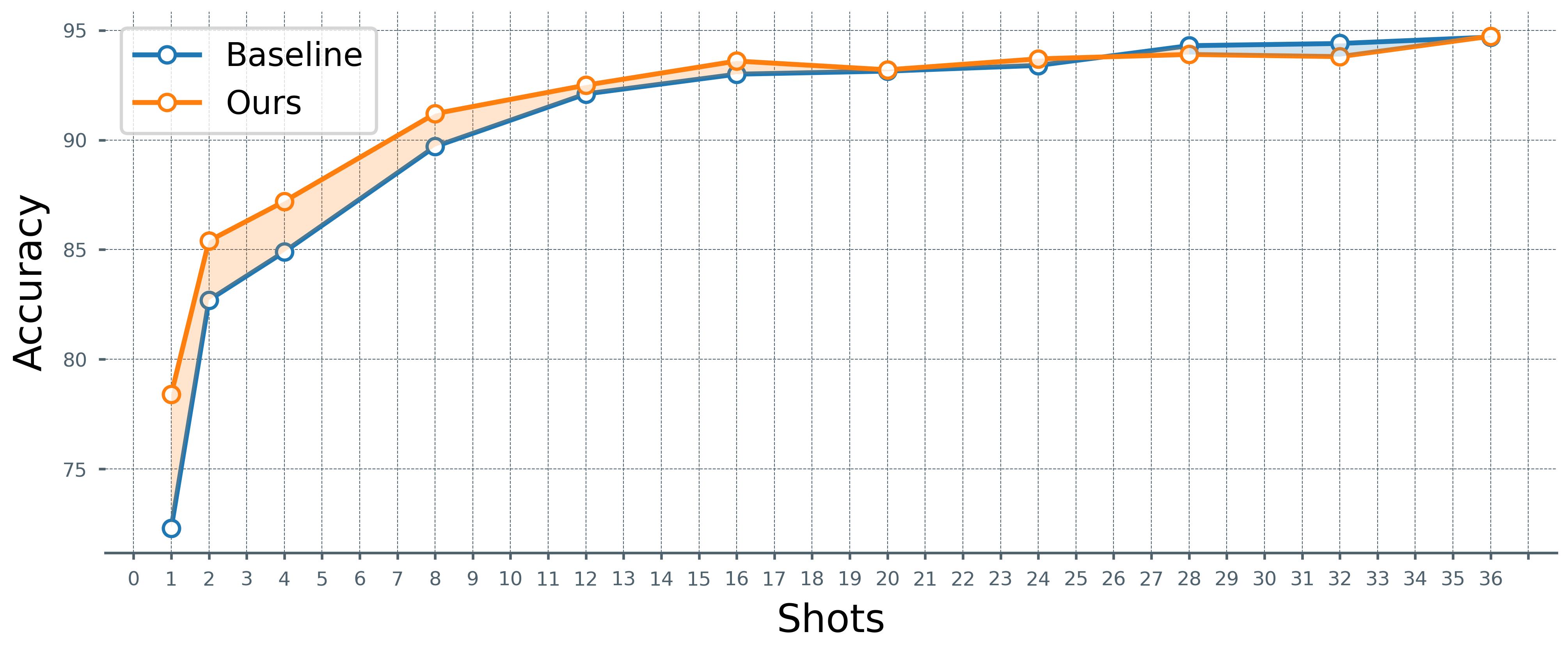}
\caption{Classification accuracy of the baseline model and our model (enhanced with the debiasing process) on the EuroSAT dataset under different shots settings.} 
\label{fig:euro_inc_shot_result}
\end{figure}
\subsection{Debiasing for Few-shot Learning}

In this work, we observe that while templates provide valuable contextual information to support classification tasks in CLIP, they also introduce biases that hinder the model's classification performance. As illustrated in Figure 2(Middle) of the main text, our analysis shows that during the fine-tuning process, the model gradually reduces its reliance on the similarity relationship between samples and templates (Template-Sample Similarity, TSS) for classification. This indicates that fine-tuning shifts the model's focus towards leveraging information directly from the samples while reducing its dependence on templates. Based on this observation, we hypothesize that as the number of labeled samples increases to a sufficient level, the model can fully utilize the information from the samples, thereby mitigating the bias introduced by templates. Consequently, our proposed debiasing method is specifically tailored for few-shot scenarios, where such biases are more pronounced.

To validate this hypothesis, we evaluate the classification accuracy of the baseline model and our proposed model (enhanced with the debiasing process) on the EuroSAT~\cite{helber2019eurosat} dataset under varying shot settings, as shown in Figure~\ref{fig:euro_inc_shot_result}. The results demonstrate that the debiasing process significantly enhances performance when the number of labeled samples is limited. However, as the number of labeled samples increases, the performance improvement diminishes. When the number of samples becomes sufficiently large, the debiasing process provides minimal to no benefit and may even slightly reduce classification accuracy (e.g., when the number of shots exceeds 28).

\section{Implementation Details}
In the model training process, we follow the setting of~\cite{zanella2024low} and use LoRA to fine-tune the model. Specifically, we apply low-rank matrices to the query, key, and value matrices with a rank ( r = 2 ) for each layer in both the vision and text encoders. We regularize the input of the LoRA module with a dropout layer, where ( p = 0.25 )~\cite{hu2021lora}. Data augmentation is performed following the method in~\cite{muller2021trivialaugment}. The learning rate is set to $2 \times 10^{-4}$ in most case, with a cosine learning rate scheduler and a batch size of 32. But in food101~\cite{bossard2014food} and ~\cite{parkhi2012cats}, a smaller learning rate of  $2 \times 10^{-5}$ is used. For each dataset, the same as~\cite{zanella2024low}, we use the template: “a photo of a {}.” to construct the prompt with the category name. During the pre-training phase, only $L_{tb}$ is used to train all parameters. The number of iterations is set to 300 times N/K (the number of labeled samples per class). In the fine-tuning phase, the loss function is the combination of the cross-entropy loss $L_{ce}$ and the debias loss $L_{tb}$:  $L_{fine} = L_{ce} + \alpha L_{tb}$ , where we set $\alpha=2$ for all datasets. The number of iterations in this phase is set to 500 times N/K (the number of labeled samples per class). A single NVIDIA GeForce RTX 3090 is used for training and testing. The overall algorithmic process is shown in Algorithm~\ref{alg:template_correction}.

\begin{algorithm}[h]
\caption{Template Correction for Few-Shot Learning}
\label{alg:template_correction}
\begin{algorithmic}[1]
\REQUIRE Pretrained CLIP model with visual encoder $\theta_v$ and text encoder $\theta_t$, set of classes $\mathcal{K}$, few-shot labeled samples $\{(\mathbf{x}_i, y_i)\}$, predefined template structures, hyperparameter $\alpha$, a set of words and phrases which mean “empty”: $T_{\text{null}}$.
\STATE \textbf{Stage1: Pretraining Initialization:}
\FOR{each labeled sample $(\mathbf{x}_i, y_i)$}
    \STATE Obtain visual embedding $\mathbf{f}_i = \theta_v(\mathbf{x}_i)$
\ENDFOR
\FOR{each word $j \in |T_{\text{null}}|$}
    \STATE Construct empty prompt $c^{E}_{j} = \text{template}(\text{word}_j)$
    \STATE Obtain empty embedding $\mathbf{t}^{E}_k = \theta_t(c^{E}_{j})$
\ENDFOR
\STATE Minimize the template bias calibration loss $L_{tb}$.
\STATE \textbf{Stage2: Few-Shot Fine-Tuning Calibration:}
\FOR{each labeled sample $(\mathbf{x}_i, y_i)$}
    \STATE Obtain visual embedding $\mathbf{f}_i = \theta_v(\mathbf{x}_i)$
\ENDFOR
\FOR{each class $k \in \mathcal{K}$}
    \STATE Construct class prompt $c_k = \text{template}(\text{class name}_k)$
    \STATE Obtain textual embedding $\mathbf{t}_k = \theta_t(c_k)$
\ENDFOR
\FOR{each word $j \in |T_{\text{null}}|$}
    \STATE Construct empty prompt $c^{E}_{j} = \text{template}(\text{word}_j)$
    \STATE Obtain empty embedding $\mathbf{t}^{E}_k = \theta_t(c^{E}_{j})$
\ENDFOR
\STATE Compute template bias calibration loss $L_{tb}$.
\STATE Compute cross-entropy loss $L_{ce}$.
\STATE Minimize the total loss $L_{fine} = L_{ce} + \alpha L_{tb}$.
\end{algorithmic}
\end{algorithm}

\section{Additional Related Work}

Prompt engineering has emerged as a critical technique for expanding the capabilities of large language models (LLMs) and visual language models (VLMs). By employing task-specific instructions, known as prompts, this approach enhances model performance without requiring modifications to the underlying parameters.
Prompt engineering was initially developed for LLMs~\cite{liu2023pre,tonmoy2024comprehensive,chen2023unleashing} and later extended to VLMs~\cite{wu2023visual,bahng2022exploring}. Early studies focused on using carefully designed textual cues~\cite{radford2019language} to guide models in achieving strong performance on downstream tasks. Subsequent advancements~\cite{brown2020language} incorporated few-shot learning, leveraging a small number of labeled examples within prompts to address more complex tasks. Further research introduced Chain-of-Thought (CoT) prompting~\cite{wei2022chain,zhang2022automatic,wang2022self,zhao2023enhancing,yao2022react,long2023large}, which fosters logical and step-by-step reasoning by decomposing problems into intermediate steps. These methodologies collectively aim to guide large models in reasoning for downstream tasks by providing rich contextual information. Despite these advancements, few studies have systematically explored the biases that the contextual information in prompts may introduce to downstream tasks.

\paragraph{Few-shot Learning}

Few-shot learning in visual classification addresses the challenge of performing classification tasks with limited labeled samples. Approaches to few-shot learning can be broadly categorized into transfer learning and meta-learning methods.
Transfer learning methods~\cite{tian2020rethinking,wang2019simpleshot,wang2020cooperative,song2023learning,zhao2023dual,wu2021selective,wu2023invariant,zhao2021graph,qiao2019transductive} leverage pre-trained models on related tasks to adapt to downstream tasks, effectively utilizing prior knowledge. Meta-learning methods, on the other hand, are typically divided into three subcategories: model-based, metric-based, and optimization-based approaches. 
Model-based methods~\cite{cai2018memory,mishra2017simple} focus on designing architectures that can quickly update parameters using a small support set to adapt to new tasks. Metric-based methods~\cite{snell2017prototypical,han2022few} emphasize modeling distance distributions between samples, ensuring that samples from the same class are closer in feature space, while those from different classes are farther apart. Finally, optimization-based methods~\cite{antoniou2018train,jamal2019task} aim to learn effective optimization strategies that enable the model to generalize using limited training data. Together, these methods form the foundation of few-shot learning and continue to be a critical area of research in visual classification.

\end{document}